%% file: main.tex
\documentclass[10pt,twocolumn,letterpaper]{article}

\usepackage{cvpr}              %

\input{preamble}
\providecommand{\celine}[1]{{\protect\color{black}{#1}}}
\definecolor{cvprblue}{rgb}{0.21,0.49,0.74}
\usepackage[pagebackref,breaklinks,colorlinks,allcolors=cvprblue]{hyperref}

\usepackage{subcaption} 
\usepackage{graphicx}
\usepackage{xcolor}
\usepackage{multirow}
\usepackage{makecell}
\usepackage{wasysym}
\usepackage{amssymb}
\usepackage{arydshln}
\usepackage[normalem]{ulem}
\usepackage{wrapfig}
\usepackage{pifont}
\usepackage{nicefrac}

\newcommand{\tabincell}[2]{\begin{tabular}{@{}#1@{}}#2\end{tabular}} %
\newcommand{\hr}[1]{\textcolor{black}{#1}}
\newcommand{\yq}[1]{\textcolor{black}{#1}}

\title{Layer- and Timestep-Adaptive Differentiable Token Compression Ratios for Efficient Diffusion Transformers}

\author{
Haoran You$^{1,2,}$\thanks{Work done while interning at Adobe.}, Connelly Barnes$^{2}$, Yuqian Zhou$^2$, Yan Kang$^2$,\\
Zhenbang Du$^1$, Wei Zhou$^1$, Lingzhi Zhang$^2$,
Yotam Nitzan$^2$, Xiaoyang Liu$^2$,\\
Zhe Lin$^2$, Eli Shechtman$^2$, Sohrab Amirghodsi$^2$, Yingyan (Celine) Lin$^1$\\[0.5em] %
$^1$Georgia Institute of Technology \hspace{1cm} $^2$Adobe Research\\
\vspace{-1.5em}
}

\begin{document}
\maketitle
\input{sec/0_abstract}    
\input{sec/1_intro}
\input{sec/2_related_work}

\input{sec/3_method}

\input{sec/4_exps}

\input{sec/5_conclusion}
{
    \small
    \bibliographystyle{ieeenat_fullname}
    \bibliography{main}
}
\input{sec/6_supple}
\end{document}

%% file: sec/0_abstract.tex
\begin{abstract}
Diffusion Transformers (DiTs) have achieved state-of-the-art (SOTA) image generation quality but suffer from high latency and memory inefficiency, making them difficult to deploy on resource-constrained devices. One major efficiency bottleneck is that existing DiTs apply equal computation across all regions of an image.
However, not all image tokens are equally important, and certain localized areas require more computation, such as objects. 
To address this, we propose \textbf{DiffCR}, a dynamic DiT inference framework with differentiable compression ratios, which automatically learns to dynamically route computation across layers and timesteps for each image token, resulting in efficient DiTs.
Specifically, DiffCR integrates three features:
(1) 
A token-level routing scheme where each DiT layer includes a router that is fine-tuned jointly with model weights to predict token importance scores. In this way, unimportant tokens bypass the entire layer's computation; %
(2) A layer-wise differentiable ratio mechanism where different DiT layers automatically learn varying compression ratios from a zero initialization, resulting in large compression ratios in redundant layers while others remain less compressed or even uncompressed;
(3) A timestep-wise differentiable ratio mechanism where each denoising timestep learns its own compression ratio. The resulting pattern shows higher ratios for noisier timesteps and lower ratios as the image becomes clearer.
Extensive experiments on text-to-image and inpainting tasks show that DiffCR effectively captures dynamism across token, layer, and timestep axes, achieving superior trade-offs between generation quality and efficiency compared to prior works.
The project website is available \href{https://www.haoranyou.com/diffcr}{here}.
\end{abstract}

%% file: sec/1_intro.tex
\vspace{-1em}
\section{Introduction}
\label{sec:intro}
\vspace{-0.2em}

Diffusion models have recently demonstrated outstanding performance in image generation, with architectures evolving from U-Nets~\cite{rombach2022high,ho2020denoising,song2020denoising,podell2023sdxl} to Transformers~\cite{peebles2023scalable,bao2023all,chen2023pixart,chen2024pixart}. Among these, Diffusion Transformers (DiTs)~\cite{bao2023all,peebles2023scalable} stand out for their superior scalability.
However, diffusion models, particularly DiTs, are hampered by substantial computational and memory demands, which limits their efficiency in generation and deployment. For instance, generating a 1024px image with full context on a single A100 GPU can take 19.48 seconds and require $>$40GB GPU memory~\cite{nitzan2024lazy,podell2023sdxl}.
\celine{One major efficiency bottleneck in most DiTs stems from the uniform application of computation across all image regions, despite varying levels of complexity in different areas~\cite{nitzan2024lazy,zeng2022not,rao2021dynamicvit}.}
Such an efficiency bottleneck suggests an ideal DiT inference framework could have adaptive and dynamic computation across three key axes in DiTs: token, layer, and timestep.

Various techniques have been proposed to address the efficiency bottleneck along the three aforementioned key axes: (1) token merging~\cite{bolya2023token}, pruning~\cite{wang2024attention}, and downsampling~\cite{smith2024todo}; (2) layer~\cite{kim2023bk} or channel~\cite{fang2023structural} pruning; and (3) few-step distillation and  sampling~\cite{salimans2022progressive,meng2023distillation,yin2024one,wang2024closer}.
While promising, the techniques (1-2) for the most part rely on heuristics, such as heuristic rules for token importance and channel and layer pruning rules. Moreover, compression ratios are often uniform across layers or adjusted empirically based on prior experience. In addition, most approaches focus on a single efficiency axis, overlooking the compounded effect of combining optimizations across all three.

\celine{
To achieve a unified and learnable dynamic DiT inference framework with differentiable compression ratios across layers and timesteps, three key challenges must be tackled: (1) \textit{Token Perspective:} Developing a learnable token importance metric that adapts to visual content, as not all tokens are equally important. (2) \textit{Layer Perspective:} Designing mechanisms to autonomously learn adaptive compression ratios for each layer, optimizing processing efficiency, since not all layers contribute equally. (3) \textit{Timestep Perspective:} Developing methods to learn and apply compression ratio patterns effectively across timesteps, as not all timesteps are equally important.
We make the following contributions to address these three challenges:}

\begin{itemize}
    \item We propose a dynamic DiT inference framework with differentiable compression ratios, dubbed \textbf{DiffCR}, which automatically learns an optimal dynamic computation across layers and timesteps for each image token, resulting in efficient DiT models for content generation tasks.

    \item \textit{\textbf{Enabler 1}}: We adopt a token-level routing scheme inspired by the mixture-of-depth (MoD)~\cite{raposo2024mixture}, which automatically learns token importance scores. Each DiT layer includes a lightweight router that is fine-tuned jointly with the model weights. Based on the compression ratio, less important tokens bypass computation in the entire layer.
    To the best of our knowledge, we are the first to apply MoD to the vision domain.
    Our routing analysis reveals that token importance varies across layers and timesteps.

    \item \textit{\textbf{Enabler 2}}: Based on our analysis, we introduce a novel DiffCR module that enables the token routing scheme to be differentiable with respect to compression ratios, allowing the model to learn adaptive compression ratios for each layer starting with zero initialization. Redundant layers learn higher compression ratios, while important layers remain less compressed or entirely uncompressed.

    \item \textit{\textbf{Enabler 3}}: We further present a timestep-wise differentiable ratio mechanism, enabling each layer and denoising timestep to learn its own compression ratio. This results in a pattern where noisier timesteps adopt higher compression ratios, while clearer stages maintain lower ratios.
\end{itemize}

Our extensive experiments on both image inpainting and text-to-image (T2I) tasks consistently demonstrate that DiffCR achieves a superior trade-off between generation quality and efficiency, with an average FID reduction of 8.51 while maintaining comparable latency and memory usage,
compared to the most competitive baseline.

%% file: sec/2_related_work.tex
\vspace{-0.2em}
\section{Related Work}
\label{sec:related_work}

\subsection{Diffusion Models}

Diffusion models~\cite{sohl2015deep,ho2020denoising} have demonstrated superior performance over prior SOTA generative adversarial networks (GANs) in image synthesis tasks~\cite{dhariwal2021diffusion}. Early diffusion models primarily utilized U-Net architectures.
Subsequent work introduced several improvements, such as advanced sampling~\cite{karras2022elucidating,song2020denoising,meng2023distillation} and classifier-free guidance~\cite{ho2022classifier}.
Although effective, these models suffered from high generation latency due to processing directly in pixel space, thus limiting their practical applications. 
The introduction of Latent Diffusion Models (LDMs)~\cite{rombach2022high} marked a significant advancement by encoding pixel space into a more compact latent space through training a Variational Auto-Encoder (VAE). This reduced the computational cost of the diffusion process, paving the way for widely used models like Stable Diffusion Models (SDMs)~\cite{podell2023sdxl}.
More recently, researchers have explored Transformer~\cite{vaswani2017attention} architectures for diffusion, leading to the development of DiTs~\cite{peebles2023scalable,bao2023all}, which employ a pure Transformer backbone and exhibit improved scalability. Our DiffCR proposes a novel dynamic DiT inference framework with differentiable compression ratios and is compatible with all recent DiT models.

\vspace{-0.1em}
\subsection{Efficient Diffusion and DiT Models}
\vspace{-0.1em}
DiTs~\cite{peebles2023scalable} are resource-intensive due to the transformer architecture, with the attention module exhibiting quadratic complexity relative to the number of tokens. Previous work has mainly focused on optimizing DiTs' deployment efficiency along three dimensions: token, layer, and timestep.
For tokens, researchers have introduced techniques like token merging~\cite{bolya2023token} to merge similar tokens, token pruning~\cite{wang2024attention} or image resolution downsampling~\cite{smith2024todo} to remove redundant tokens, and LazyDiffusion~\cite{nitzan2024lazy}, which is specialized for the inpainting task and bypasses generating background tokens. For layers, methods such as layer~\cite{kim2023bk} and channel~\cite{fang2023structural} pruning, as well as intermediate feature caching~\cite{xu2018deepcache,moura2019cache,ma2024learning}, have been proposed to skip redundant computations. For timesteps, strategies include distillation to reduce the required number of timesteps, which has been explored for UNets~\cite{salimans2022progressive,meng2023distillation,yin2024one,kang2024distilling,yin2024improved,sauer2025adversarial} although there is no reason to believe these techniques cannot apply to Transformers, and asymmetric sampling, which has been applied to Transformer architectures and allocates more samples to undersampled stages and fewer to stages that have already converged~\cite{wang2024closer,pu2024efficient}.
Additionally, to accelerate diffusion T2I models, more specialized techniques have been introduced~\cite{chen2023pixart,chen2024pixart}.
In contrast, our proposed DiffCR is a learnable and unified dynamic DiT inference framework with differentiable compression ratios across layers and timesteps, exploring the compounded effects of compression across all three axes. However, we do not explore few step distillation (e.g.~\cite{yin2024improved}) in this paper, since it is an orthogonal acceleration method that is complementary to ours.

\begin{figure*}[t]
    \centering
    \includegraphics[width=\linewidth]{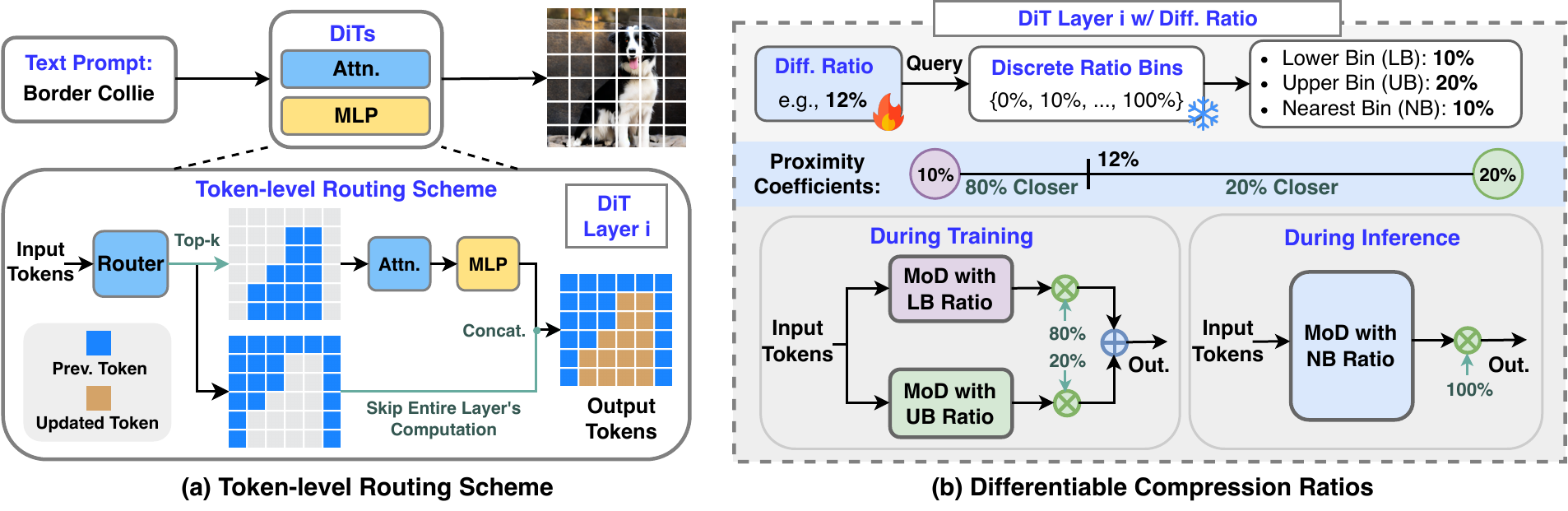}
    \vspace{-1.5em}
    \caption{Overview of the proposed DiffCR framework: (a) token-level routing scheme and (b) differentiable compression ratios. 
    }
    \vspace{-1em}
    \label{fig:overview}
\end{figure*}

\vspace{-0.1em}
\subsection{Dynamic Inference}
\vspace{-0.1em}
Model compression~\cite{deng2020model} offers a static approach to improving inference efficiency, while dynamic inference~\cite{wang2020dual,zhao2024dynamic,wu2018blockdrop,wang2018skipnet,raposo2024mixture} enables adaptive compression based on the input, layer, or other conditions. 
For example, early exiting methods~\cite{teerapittayanon2016branchynet,huang2017multi,moon2023early} predict the optimal point for early termination within intermediate layers, allowing the model to exit before completing all computations.
Dynamic layer-skipping methods~\cite{wang2020dual,wang2018skipnet,wu2018blockdrop} selectively execute subsets of layers for each input, often utilizing a gating network to make decisions on the fly. At a finer granularity, researchers have also explored channel skipping~\cite{mullapudi2018hydranets,fang2023structural} and mixture-of-depths (MoD) approaches~\cite{raposo2024mixture}, which select specific subsets of layers for individual tokens rather than processing the entire input uniformly.
In contrast, our DiffCR is the first to introduce a unified dynamic DiT inference framework that optimizes across three axes: token, layer, and timestep. It also enables differentiable compression ratios that are fine-tuned jointly with the network, enhancing adaptability and efficiency.

%% file: sec/3_method.tex
\section{The Proposed DiffCR Framework}
\label{sec:methods}

In this section, we present the proposed DiffCR framework. First, we provide an overview of the method. Then, we detail the three enablers: (1) the token-level routing scheme for DiTs in Sec.~\ref{sec:enabler-1}; (2) the layer-wise differentiable compression ratio scheme in Sec.~\ref{sec:enabler-2}; and (3) the timestep-wise differentiable compression ratio scheme in Sec.~\ref{sec:enabler-3}.

\subsection{Overview of DiffCR}

Motivated by the need for unified and dynamic compression during DiT inference, DiffCR introduces a token-level routing scheme to dynamically learn the importance of each token on the fly.
As illustrated in Fig.~\ref{fig:overview} (a), similar to previous mixture-of-depths (MoD) work~\cite{raposo2024mixture} for NLP tasks, each DiT layer incorporates a lightweight router using a single linear layer to predict the importance of each token based on the input image/noise and text embedding. This allows us to bypass computations for less important tokens in each layer and to directly forward their activations to the layer's outputs.
Consequently, each token is processed by only a selective subset of layers.
Visualization of these routers' predictions reveals that different layers or timesteps favor varying compression ratios—for instance, some layers prioritize generating objects, while others focus on backgrounds—highlighting the need for adaptive compression across layers and timesteps.
To achieve such dynamic compression, DiffCR incorporates a differentiable compression ratio scheme, as shown in Fig.~\ref{fig:overview} (b). This scheme includes a learnable scalar parameter that represents a continuous compression ratio, and predefined discrete ratio bins as proxy ratios. The scalar queries the bins to identify lower and upper bin ratios, creating two separate paths with distinct compression ratios. The final output is a linear combination of these two paths, weighted by the proximity of the learned ratio to each bin. We apply a mean-squared error (MSE) loss to ensure that the average learned ratio across layers or timesteps converges to the target ratio.
By doing so, DiffCR learns the adaptive compression ratios in a differentiable manner, resulting in efficient and dynamic mixture-of-depths DiTs.

\subsection{Enabler 1: Token-level Routing Scheme}
\label{sec:enabler-1}

\textbf{Motivation.}
We are motivated by the varying computational demands across tokens, as many of them require fewer layers for efficient processing. We start from the same token-level routing scheme as MoD~\cite{raposo2024mixture}. We remove from MoD two features that were specialized for the acausal NLP task: specifically, we remove the auxiliary loss and auxiliary MLP predictor from Section 3.5 of their paper. To the best of our knowledge, our paper is the first application of MoD to the vision domain, so we next review the routing mechanism, perform some visualizations, and report insights for vision tasks.   %

\begin{figure*}
    \centering
    \includegraphics[width=\linewidth]{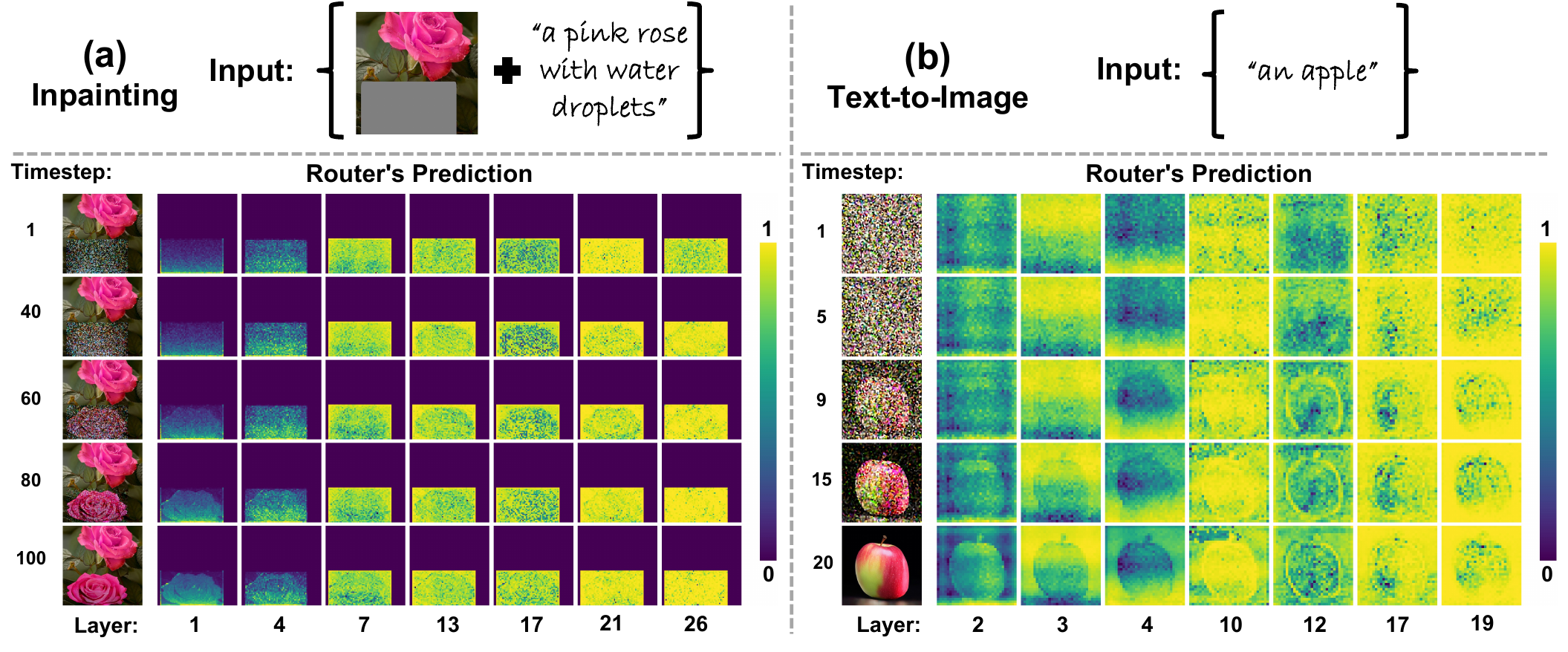}
    \vspace{-2em}
    \caption{Visualization of the router's predictions: (a) For inpainting tasks, where inputs are masked images with text prompts, we follow the previous SOTA method Lazy-Diffusion~\cite{nitzan2024lazy} to generate only the masked area rather than the entire image; (b) For text-to-image (T2I) tasks, where inputs are noise and text prompts, we follow PixArt-$\Sigma$~\cite{chen2024pixart} for generation. 
    Each visualization includes the router's prediction map with values ranging from 0 to 1. The generated image at each corresponding timestep is shown on the left, while the router's prediction maps across various layers and timesteps are displayed on the right.
    More visualizations are provided in the supplementary materials.
    }
    \vspace{-1.3em}
    \label{fig:router_visualization}
\end{figure*}

\begin{figure}[t]
    \centering
    \includegraphics[width=\linewidth]{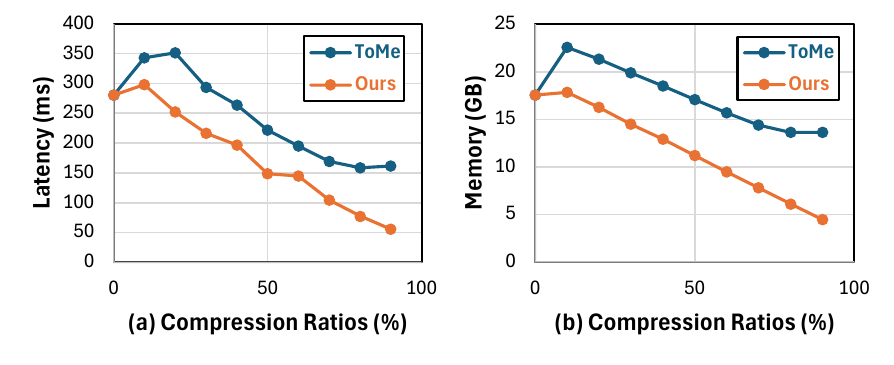}
    \vspace{-1.5em}
    \caption{Comparison of latency and memory savings between our DiffCR router and the previous token merging method (ToMe)~\cite{bolya2023token} when applied to ViT-XL/2~\cite{vit,nitzan2024lazy} on an A100 GPU.}
    \vspace{-2em}
    \label{fig:tome_vs_mod}
\end{figure}

\textbf{Token-level Routing.}
DiTs process noise and conditional text embeddings as inputs, aiming to denoise and generate images in an end-to-end manner.
To predict token importance, we employ a simple yet effective token-level routing scheme from MoD~\cite{raposo2024mixture}. As illustrated in Fig.~\ref{fig:overview} (a), each DiT layer incorporates a lightweight router composed of a single linear layer with a sigmoid activation function, predicting each token's importance on a scale from 0 to 1. %
After passing through the routers, we select the top-$k$ most important tokens for this layer's processing, while the activations of other tokens are cached and concatenated with the layer outputs, bypassing the entire layer computation, including both attention and MLPs. 
To enable gradient flow to the router’s weights during joint fine-tuning with pretrained DiT models, the same as MoD~\cite{raposo2024mixture}, we rescale the top-$k$ token output activations by multiplying them with the router’s predictions. This rescaling ensures that gradients are propagated effectively to the router during backpropagation.
The value of $k$ is determined based on compression ratios; we empirically find that 20\% or 30\% compression offers an optimal trade-off between latency/memory efficiency and minimal degradation in generation quality. Unlike previous token merging techniques~\cite{bolya2023token}, where computational savings are not directly proportional to the merge ratios due to the extra overhead, as shown in Fig.~\ref{fig:tome_vs_mod}, the MoD in our approach yields more reductions in actual latency and memory usage due to negligible overhead.

\textbf{Router Visualization and Insights.}
To test the router's effectiveness, we visualize its predictions in Fig.~\ref{fig:router_visualization}.
Our observations reveal that:
(1) \textit{The router effectively captures semantic information}, clearly delineating object shapes and achieving an attention-like effect with significantly reduced computational costs;
(2) \textit{The predicted token importance varies across layers and timesteps}. For instance, some layers prioritize object generation, while others emphasize background areas. Additionally, as timesteps progress, the router increasingly captures the semantic contours of objects, underscoring the need for dynamic token importance estimation;
(3) \textit{\textit{The optimal compression ratio differs across layers and timesteps}}. For example, certain layers designate all tokens as high-importance, showing minimal redundancy, whereas other layers selectively prune object or background tokens with distinct shapes, requiring varying compression ratios. Similar variance is observed across timesteps. \yq{In the current MoD, a fixed global compression rate is applied equally to each layer and timestep, rather than adapting to its individual significance}. Uniform pruning risks over-pruning critical layers or timesteps while leaving redundant ones less compressed. This motivates us to apply adaptive and dynamic compression ratios across both layers and timesteps. 

\subsection{Enabler 2: Layer-wise Differentiable Ratio}
\label{sec:enabler-2}

\textbf{Motivation.}
Recognizing that different layers prioritize different objects or background elements and thus benefit from distinct compression ratios, we propose a novel layer-wise differentiable compression ratio mechanism. This approach automatically learns each layer's compression ratio from a zero initialization in a differentiable manner, adapting to the varying redundancy levels across layers.

\textbf{Design Choice.}
Before designing DiffCR, we address a key choice: a discrete proxy or a continuous ratio representation. Previous work~\cite{chen2023diffrate} uses a discrete proxy with multiple compression ratio candidates and learnable probabilities, but this approach poses three challenges for MoD: (1) it lacks effective initialization, as the final ratio relies on the product of candidates and probabilities, making it difficult to initialize all ratios at zero; (2) MoD requires a differentiable and learnable router, incompatible with discrete proxies that need multiple sets of top-$k$ tokens; and (3) it introduces numerous learnable parameters, complicating training and interpretability, and leading to hard-to-interpret ratio distributions. In contrast, we represent each layer’s compression ratio with a single continuous scalar parameter, reducing the number of parameters to 28 for DiT’s 28 layers~\cite{chen2023pixart} and directly representing the compression ratio.

\begin{figure}[t]
    \centering
    \includegraphics[width=\linewidth]{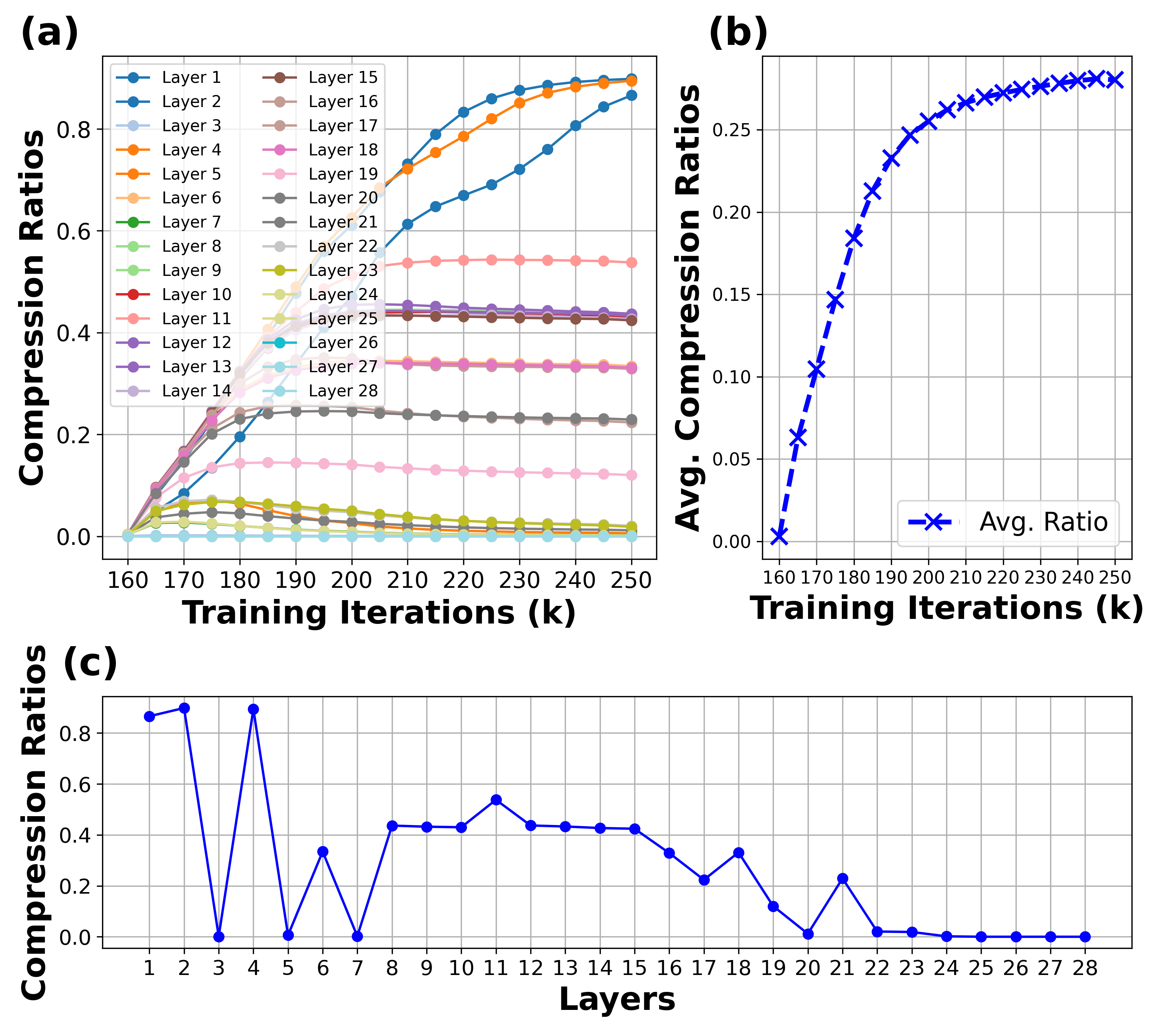}
    \vspace{-2em}
    \caption{Visualization of the 
    compression ratio
    trajectory during fine-tuning: (a) Trajectories for each of the 28 layers in DiT models; (b) Average ratio trajectory across all layers; and (c) The final learned ratio distribution across 28 layers.}
    \vspace{-1.5em}
    \label{fig:layerwise_trajectory}
\end{figure}

\textbf{Layer-wise DiffCR.}
As illustrated in Fig.~\ref{fig:overview} (b), we assign each layer a single learnable parameter and introduce discrete MoD compression ratio bins at 10\% intervals, ranging from 0\% to 100\%.
During training, the learnable MoD ratio queries the nearest two discrete bins to retrieve the lower and upper bin ratios. For example, a 22\% learnable ratio would correspond to 20\% as the lower bin and 30\% as the upper bin. We then apply a forward pass through the DiT layer (with MoD routers) with each of these bin ratios, producing two output branches. The final output is a weighted linear combination of these branches, where the weights are determined by the proximity of the learnable ratio to each bin—e.g., for 12\%, the output would be 80\% weighted towards the 10\% branch and 20\% towards the 20\% branch.
Although this approach doubles the cost of a forward pass during training, we simply select the nearest bin as the final compression ratio during inference, eliminating any overhead.
To ensure that the ratio converges to our target value, we incorporate an additional MSE loss between the current learned average ratios across all layers in the batch and the target ratio, which is a hyperparameter.

\textbf{Ratio Trajectory Analysis.}
We visualize the training trajectory of compression ratios for all layers during fine-tuning of a LazyDiffusion model on an inpainting task in Fig.~\ref{fig:layerwise_trajectory} (a-c).
The visualization reveals that: 
(1) Each layer learns its unique compression ratio, with redundant layers achieving higher compression and critical layers remaining less or entirely uncompressed;
(2) The average ratio across layers gradually converges to the target ratio. In this example, with a target of 30\%, the final achieved average ratio is approximately 29\%, indicating a minor gap. Notably, a trade-off exists between convergence speed and generation quality: a higher MSE loss coefficient for the ratio accelerates convergence but may degrade quality due to overly rapid compression, while a smaller coefficient promotes gradual convergence and maintains quality, albeit with slower training. In practice, we set the coefficient to 0.3 to balance speed and quality effectively;
(3) The middle layers exhibit greater redundancy, while the later layers generally have less redundancy and often cannot be compressed. The early layers show variable redundancy levels.

\begin{figure}[t]
    \centering
    \includegraphics[width=\linewidth]{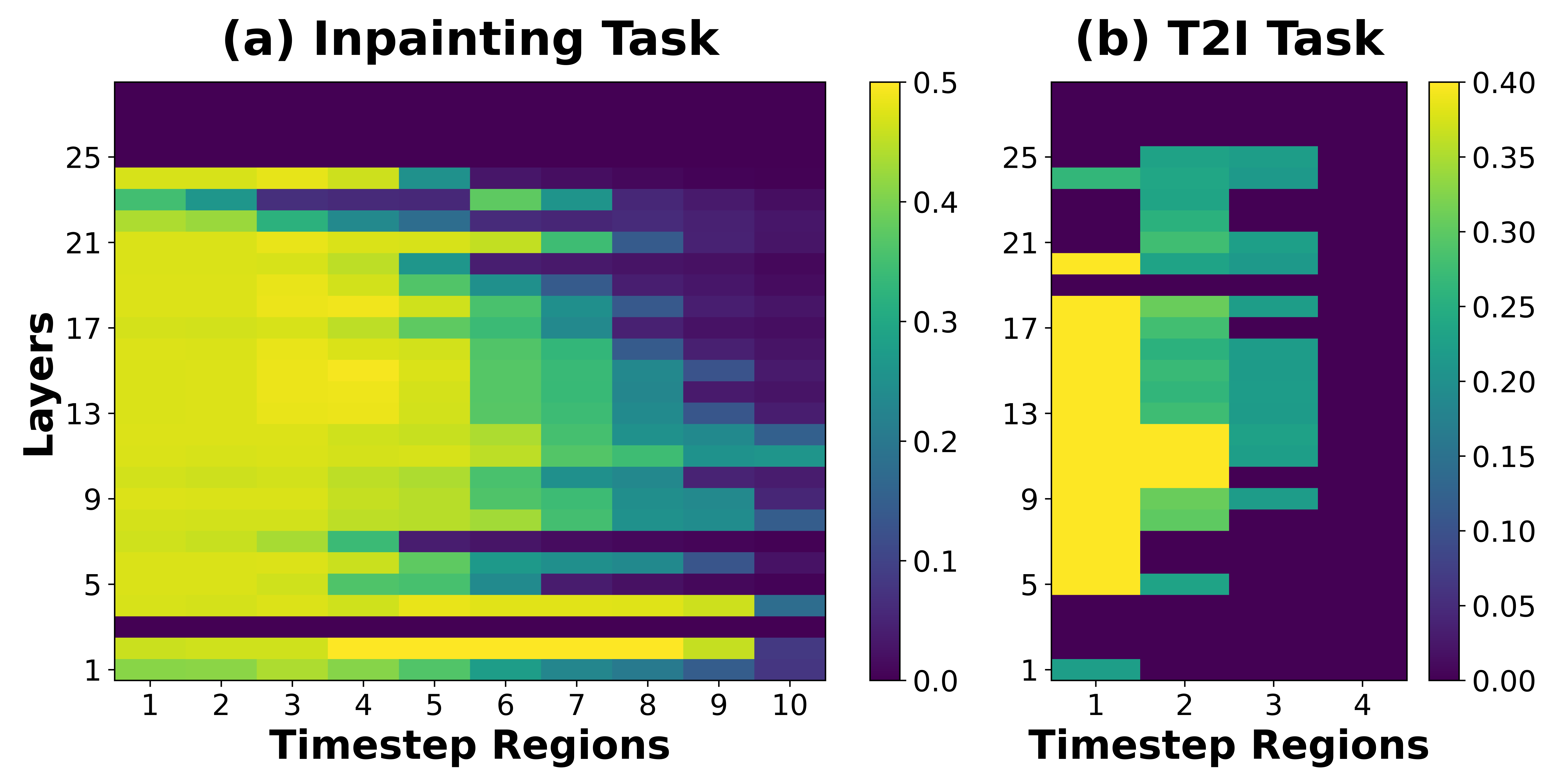}
    \vspace{-2em}
    \caption{Visualization of the learned ratio patterns across both timesteps and layers for the (a) inpainting task and (b) T2I task. 
    }
    \vspace{-1.5em}
    \label{fig:timestep_wise}
\end{figure}

\subsection{Enabler 3: Timestep-wise Differentiable Ratio}
\label{sec:enabler-3}

\textbf{Motivation.}
In addition to layer-wise ratio variances, we also observe that the model exhibits varying levels of redundancy across timesteps. This motivates us to explore an approach for timestep-wise compression ratios as well.

\textbf{Timestep-wise Differentiable Ratio.}
On top of the layer-wise DiffCR, we introduce learnable parameters specific to different timestep regions. 
For the image inpainting task, following the previous SOTA LazyDiffusion~\cite{nitzan2024lazy}, we use 1,000 training timesteps and 100 sampling timesteps, which we evenly divide into 10 regions. We assign 10 learnable ratios per layer accordingly, resulting in a total of 280 learnable parameters.
Similarly, for T2I tasks, following PixArt-$\Sigma$~\cite{chen2024pixart} with 20 timesteps, we divide them into 4 regions, assigning 4 learnable ratios per layer, yielding 112 learnable parameters.
The same as before, we apply an MSE loss between the averaged learned ratios within the batch and the target ratio to ensure convergence.

\textbf{Ratio Pattern Analysis.}
We visualize the learned compression ratio patterns across both timesteps and layers in Fig.~\ref{fig:timestep_wise}.
For both image inpainting and T2I tasks, we consistently observe that noisy timesteps (corresponding to earlier sampling timesteps or later training timesteps) exhibit higher redundancy and allow for higher compression, whereas timesteps where images become clearer (corresponding to later sampling timesteps or earlier training timesteps) show less redundancy. This learned pattern aligns with previous empirical findings~\cite{wang2024closer}, which suggest that high-noise timesteps are associated with convergence regions containing easier samples, allowing for fewer sampling timesteps, while low-noise timesteps involve harder samples and require more frequent sampling.

%% file: sec/4_exps.tex
\vspace{-0.2em}
\section{Experiments}
\label{sec:exps}
\vspace{-0.2em}
\subsection{Experiment Settings}
\label{sec:exp_setting}

\textbf{Tasks, Datasets, and Models.}
\uline{\textit{Tasks \& Datasets.}} We evaluate our DiffCR on two representative image generation tasks using corresponding benchmark datasets: (1) an image inpainting task on an internal dataset of 220 million high-quality images, covering diverse objects and scenes. Masks and text prompts are generated following~\cite{nitzan2024lazy,xie2023smartbrush}; and (2) a T2I task on the LAION-5B dataset~\cite{schuhmann2022laion}, restricted to image samples with high aesthetic scores, English text, and a minimum text similarity score of 0.24.
\underline{\textit{Models.}} 
We integrate our proposed DiffCR approach with SOTA models. For the inpainting task, we use Lazy Diffusion (an adapted PixArt-$\alpha$ model with an additional ViT encoder) to generate images at 1024$\times$1024 resolution. For the T2I task, we use PixArt-$\Sigma$ to generate images at 512$\times$512 resolution.

\textbf{Training and Sampling Setting.} 
For the inpainting task, we fine-tune the model parameters until convergence using the AdamW optimizer~\cite{ilya2019adamw} with a learning rate of $10^{-4}$ and weight decay of $3\times 10^{-2}$. For sampling, images are generated using IDDPM~\cite{pmlr-v139-nichol21a} with 100 timesteps and a CFG factor of 4.5.
For the T2I task, we fine-tune the model using a LoRA adapter with a rank of 32 until both training and validation losses converge, and the MSE loss between the current and target compression ratios drops to approximately zero for DiffCR models. During training, we calculate the diffusion loss with IDDPM~\cite{pmlr-v139-nichol21a} over 1K timesteps. For sampling, we generate images using DPM-solver~\cite{dpm-solver} with 20 timesteps and a CFG factor of 4.5.
All training is conducted on a cluster of 8$\times$A100-80GB GPUs.

\textbf{Baselines and Evaluation Metrics.}
\uline{\textit{Baselines.}} For both the T2I and inpainting tasks, we compare the proposed DiffCR against SOTA baselines, including ToMe~\cite{bolya2023token}, AT-EDM~\cite{wang2024attention}, and our adapted MoD with uniform MoD compression ratio. For the inpainting task, we also compare against \textit{RegenerateCrop}, which generates a tight square crop around the masked region, similar to popular software frameworks~\cite{stable_diffusion_web,diffusers}, and \textit{RegenerateImage}, which generates the entire image, as commonly done in the literature~\cite{podell2023sdxl,rombach2022high,xie2023smartbrush,wang2023imagen}.
\uline{\textit{Evaluation Metrics.}} 
We assess the generated image quality using FID scores~\cite{fid}, text-image alignment using CLIP scores~\cite{hessel2021clipscore}, and efficiency through inference FLOPs, latency, and memory usage, all measured on an A100 GPU.
For inpainting and T2I models, we evaluate on 10K images from LAION-400M~\cite{schuhmann2021laion} or LAION-5B~\cite{schuhmann2022laion}, excluding training samples, respectively.

\subsection{DiffCR over SOTA Baselines}
\label{sec:exps_overall}
\vspace{-0.2em}
\input{tabs/t2i}
\input{tabs/in-painting}

\textbf{Text-to-Image.}
To assess the effectiveness of our proposed DiffCR, we apply our proposed DiffCR to the general text-to-image task and compare it with previous token merging~\cite{bolya2023token} and pruning~\cite{wang2024attention} baselines. Specifically, we apply these compression methods on PixArt-$\Sigma$, a SOTA publicly accessible T2I model known for its high-resolution image generation quality and efficiency tradeoffs.
As shown in Tab.~\ref{tab:t2i}, PixArt-$\Sigma$ with DiffCR significantly improves generation quality, achieving \hr{57.83 and 241.11} FID reductions over ToMe~\cite{bolya2023token} and AT-EDM~\cite{wang2024attention}, respectively, with comparable or even lower latency \hr{($\downarrow$8.59\%$\sim$20.15\%)} and memory usage \hr{($\downarrow$-2.71\%$\sim$0.72\%)}. \hr{Also, under similar latency compared to ToMe~\cite{bolya2023token} with 20\% compression ratio, DiffCR achieves 335.23 FID reductions.}
Moreover, PixArt-$\Sigma$ with DiffCR also achieves comparable image generation quality to uncompressed PixArt-$\Sigma$, while delivering \hr{20.68\% and 8.33\%} latency and memory savings. Note that we compare with fine-tuned PixArt-$\Sigma$ on the LAION datasets for a fair comparison.
This set of experiments demonstrates the effectiveness of DiffCR for general T2I tasks.

\textbf{Image Inpainting.}
We further extend DiffCR to the inpainting task. Specifically, we apply it on top of the SOTA Lazy Diffusion (LD)~\cite{nitzan2024lazy}, which uses a DiT decoder to generate only the masked areas rather than the entire image, leveraging a separate ViT encoder to capture the global context of the input masked images. We compare our DiffCR approach against two types of baselines: (1) RegenerateImage and RegenerateCrop, and (2) LD with previous token merging~\cite{bolya2023token} or pruning~\cite{wang2024attention} techniques.
As shown in Tab.~\ref{tab:in-painting}, our DiffCR consistently outperforms all baselines in terms of accuracy-efficiency tradeoffs. For example, LD with DiffCR achieves FID reductions of \hr{47.35 and 189.93} compared to LD with ToME~\cite{bolya2023token} or AT-EDM~\cite{wang2024attention}, while achieving similar or up to \hr{23.61\% and 13.63\%} higher latency and memory savings. \hr{Also, under similar memory usage compared to ToMe \cite{bolya2023token} with 30\% compression ratio, LD with DiffCR achieve 265.94 FID reduction while delivering up to 21.54\% latency savings.}
Moreover, compared to RegenerateImage, our method achieves \hr{73.51\%/60.26\%} FLOPs and latency savings when inpainting 256$^2$ mask sizes within 1024$^2$ images. Notably, like Lazy Diffusion, our method’s complexity scales with mask size, while RegenerateImage generates based on full image resolution, making it less efficient for smaller mask sizes.
In comparison to RegenerateCrop, our method achieves significantly higher image generation quality \hr{($+$41.01 FID)} while also delivering 22.96\% memory savings. 
Note that all memory measurements are taken with a batch size of 128.
This set of experiments validates the effectiveness of our DiffCR when applied to image inpainting tasks.

\begin{figure*}
    \centering
    \includegraphics[width=\linewidth]{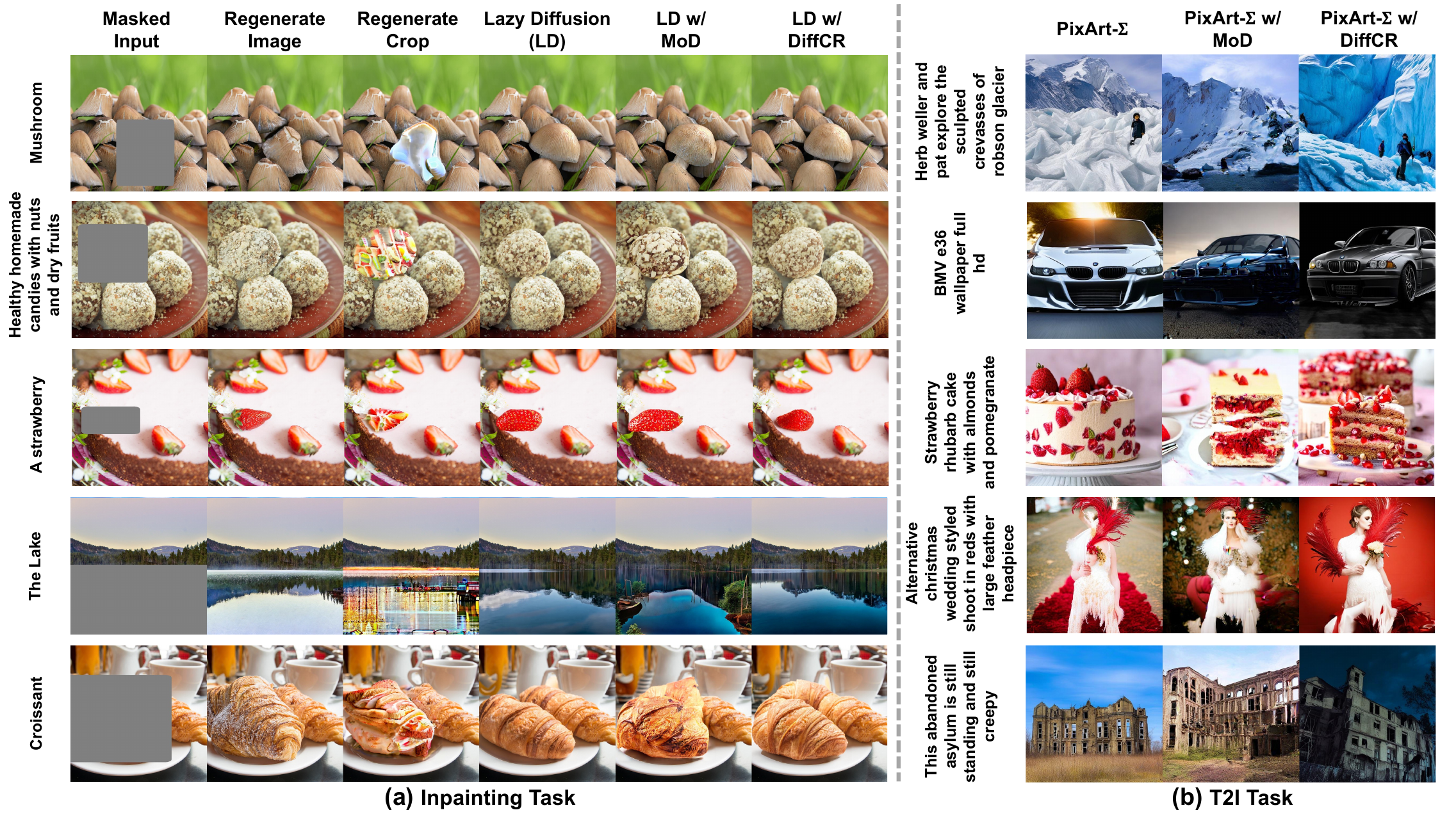}
    \vspace{-1.8em}
    \caption{Visual comparisons of our DiffCR with previous uncompressed models and SOTA compression methods: (a) Inpainting tasks, where DiffCR is applied to the LD models~\cite{nitzan2024lazy}, and (b) T2I tasks, where DiffCR is applied to PixArt-$\Sigma$~\cite{chen2024pixart}.}
    \vspace{-1em}
    \label{fig:visual_examples}
\end{figure*}

\subsection{Ablation Studies of DiffCR}
\label{sec:exps_ablation}
\vspace{-0.2em}
We conduct ablation studies on DiffCR, analyzing the contributions of the three enablers described in Sec.~\ref{sec:methods}.
As shown in Tabs.~\ref{tab:t2i} and \ref{tab:in-painting}, we report the performance of LD or PixArt-$\Sigma$ with MoD routers (Sec.~\ref{sec:enabler-1}), DiffCR-L (Sec.~\ref{sec:enabler-2}), DiffCR-LT (Sec.~\ref{sec:enabler-3}) for T2I and inpainting tasks, respectively.
The results consistently show that all components of our DiffCR contribute to the final performance. Specifically, MoD alone achieves average FID reductions of \hr{323.13 and 229.01} compared to ToMe~\cite{bolya2023token} and AT-EDM~\cite{wang2024attention} for the T2I task, with comparable or even lower latency and memory usage. DiffCR-L and DiffCR-LT further enhance the generation quality, achieving additional FID reductions of \hr{4.81/4.92 for the inpainting task and 10.5/12.1 for the T2I task}.

\begin{wrapfigure}{r}{0.47\linewidth}
    \centering
    \vspace{-2em}
    \includegraphics[width=\linewidth]{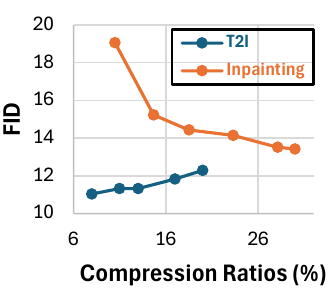}
    \vspace{-2em}
    \caption{Model trajectories.
    }
    \label{fig:model_trajectory}
    \vspace{-1.5em}
\end{wrapfigure}
Also, a key benefit of our DiffCR is that during fine-tuning, the average compression ratios across all layers gradually converge to the target ratio, producing a series of ``by-product'' models with a range of compression ratios. 
As shown in Fig.~\ref{fig:model_trajectory}, 
we visualize the model trajectory with corresponding FID scores and compression ratios for both inpainting and T2I tasks. The observations indicate that, for T2I, the FID gradually increases with compression ratio, achieving the desired results. In contrast, for inpainting, the FID gradually decreases. This difference arises because, compared to T2I, inpainting tasks and LD models are more sensitive to pruning and require longer fine-tuning to boost the generation quality. This is also reflected in Tabs.~\ref{tab:in-painting} and \ref{tab:t2i}, where FID increases post-pruning for inpainting, while it even decreases for T2I.

\subsection{Qualitative Visual Examples}
\label{sec:exps_visualization}

\textbf{Visual Examples.}
We select challenging input prompts to evaluate the qualitative results of our proposed DiffCR.
As shown in Fig.~\ref{fig:visual_examples}, the examples demonstrate that DiffCR achieves comparable or even superior generation quality compared to the RegenerateCrop baseline and even uncompressed LD or PixArt-$\Sigma$ for inpainting and T2I tasks, respectively. Note that ToMe and AT-EDM are omitted here due to their poor generation quality when applied to DiTs, even at a mere 10\% compression ratio.

\textbf{Human Preference Scores.}
We use a computer vision model to estimate likely human preferences and assess the models' ability to generate high-quality, contextually relevant images. Specifically, we generated 2K samples for the T2I task and used HPSv2~\cite{wu2023human} to evaluate human preferences for images generated by different methods.
As shown in Tab.~\ref{tab:hps}, for T2I, we apply all compression methods to PixArt-$\Sigma$~\cite{chen2024pixart}. DiffCR achieves a higher human preference score of 4.685/0.847 compared to previous compression methods, ToMe~\cite{bolya2023token} and vanilla MoD~\cite{raposo2024mixture}, respectively.

\input{tabs/hps}

%% file: tabs/t2i.tex
\begin{table*}[t]
  \centering
  \setlength{\tabcolsep}{3.5pt}
  \renewcommand{\arraystretch}{1.2}
  \caption{Quantitative comparison of DiffCR with other baselines on the T2I task. All experiments are fine-tuned from the pre-trained PixArt-$\Sigma$~\cite{chen2024pixart} on the LAION-5B~\cite{schuhmann2022laion} dataset. C.R. denotes compression ratios. We report FID ($\downarrow$) and CLIP Score ($\uparrow$) on 10K images (excluding training samples) as quality metrics and measure FLOPs ($\downarrow$), latency ($\downarrow$), and memory ($\downarrow$) on an A100 GPU as efficiency metrics under batch sizes of 16 and 128. Memory is averaged across all layers. ``-L'' and ``-LT'' indicate layer-wise or layer/timestep-wise DiffCR. ``TF'' denotes training-free methods.}
  \vspace{-0.5em}
  \resizebox{\linewidth}{!}{
        \begin{tabular}{l|c|cc|ccc|ccc}
        \Xhline{3\arrayrulewidth}
        \multicolumn{1}{l|}{\multirow{2}[2]{*}{\textbf{Methods}}} & \multicolumn{1}{c|}{\multirow{2}[2]{*}{\textbf{DiT C.R.}}} & \multicolumn{2}{c|}{\textbf{Quality}} & \multicolumn{3}{c|}{\textbf{DiT Efficiency (512$^2$; BS = 16)}} & \multicolumn{3}{c}{\textbf{DiT Efficiency (512$^2$; BS = 128)}} \\
        \cline{3-10}  &   & \textbf{FID} & \textbf{CLIP Score} & \textbf{FLOPs (G)} & \multicolumn{1}{c}{\textbf{Lat. (s)}} & \multicolumn{1}{c|}{\textbf{Mem. (GB)}} & \textbf{FLOPs (G)} & \multicolumn{1}{c}{\textbf{Lat. (s)}} & \multicolumn{1}{c}{\textbf{Mem. (GB)}} \\
        \Xhline{3\arrayrulewidth}
        PixArt-$\Sigma$~\cite{chen2024pixart} & 0\% & 151.0 & 0.173  &  17361.7  & 225.38 & 1.798 & 138893.9 & 1784.64 &	13.107\\
        PixArt-$\Sigma$ (Fine-tuned) & 0\% & 11.93  &  0.242 & 17361.7  &  225.38 & 1.798 & 138893.9 & 1784.64 & 13.107\\
        PixArt-$\Sigma$ w/ ToMe (TF)~\cite{bolya2023token} & 10\% & 68.51  & 0.211  &  16391.4 & 224.04 & 1.674 & 131131.6 & 1772.84	& 12.087\\
        PixArt-$\Sigma$ w/ ToMe (TF)~\cite{bolya2023token} & 20\% & 345.91  & 0.122 & 15421.2 & 213.79 & 1.546 & 123369.2 & 1681.83 &  11.078\\
        PixArt-$\Sigma$ w/ AT-EDM (TF)~\cite{wang2024attention} & 20\% & 251.79  & 0.129  & 15132.7  & 196.88  & 1.617  & 121061.2  & 1548.80  & 11.870 \\
        \Xhline{3\arrayrulewidth}
        PixArt-$\Sigma$ w/ \textbf{MoD} & 20\% & 22.78  & 0.207  & 13949.3 &  178.89 & 1.659 & 111594.2 & 1402.96	& 11.987 \\
        PixArt-$\Sigma$ w/ \textbf{DiffCR-L} & 20\% & 12.28  &  0.232 &  13967.1 & 180.84  & 1.662 & 111737.0 & 1425.30 & 12.015\\
        PixArt-$\Sigma$ w/ \textbf{DiffCR-LT} & 20\% &  10.68 &  0.238 & 13957.2 & 179.71  & 1.664 & 111657.3	& 1415.63 & 12.021\\
        \Xhline{3\arrayrulewidth}
        \end{tabular}%

  }
  \label{tab:t2i}%
\end{table*}%

%% file: tabs/in-painting.tex
\begin{table*}[t]
  \centering
  \setlength{\tabcolsep}{3.5pt}
  \renewcommand{\arraystretch}{1.2}
  \caption{Quantitative comparison of DiffCR with other baselines on the inpainting task. Scores for SDXL~\cite{podell2023sdxl} are provided for reference only and are not directly comparable. C.R. denotes compression ratios. We report FID ($\downarrow$)~\cite{fid} and CLIP Score ($\uparrow$)~\cite{hessel2021clipscore} on 10K images from LAION-400M~\cite{schuhmann2021laion} as quality metrics, and measure FLOPs ($\downarrow$), latency ($\downarrow$), and memory usage ($\downarrow$) on an A100 GPU as efficiency metrics for two inpainting mask sizes (512$^2$ and 256$^2$ within 1024$^2$ images). ``-L'' and ``-LT'' indicate layer-wise or layer/timestep-wise DiffCR. ``TF'' denotes training-free methods.}
  \vspace{-0.5em}
  \resizebox{\linewidth}{!}{
        \begin{tabular}{l|cc|cc|ccc|ccc}
        \Xhline{3\arrayrulewidth}
        \multicolumn{1}{l|}{\multirow{1}[4]{*}{\textbf{Methods}}} & \multicolumn{1}{c}{\multirow{1}[4]{*}{\tabincell{c}{\textbf{ViT En.}\\\textbf{C.R.}}}} & \multicolumn{1}{c|}{\multirow{1}[4]{*}{\tabincell{c}{\textbf{DiT De.}\\\textbf{C.R.}}}} & \multicolumn{2}{c|}{\textbf{Quality}} & \multicolumn{3}{c|}{\textbf{DiT Efficiency (512$^2$ within 1024$^2$)}} & \multicolumn{3}{c}{\textbf{DiT Efficiency (256$^2$ within 1024$^2$)}} \\
        \cline{4-11}  &   &   & \textbf{FID} & \textbf{CLIP Score} & \textbf{FLOPs (G)} & \multicolumn{1}{c}{\textbf{Lat. (s)}} & \multicolumn{1}{c|}{\textbf{Mem. (GB)}} & \multicolumn{1}{c}{\textbf{FLOPs (G)}} & \multicolumn{1}{c}{\textbf{Lat. (s)}} & \multicolumn{1}{c}{\textbf{Mem. (GB)}} \\
        \Xhline{3\arrayrulewidth}
        SDXL~\cite{podell2023sdxl} & N/A & 0\% & 6.37  & 0.2112  & 5979.5   & 66.09  & OOM  & 5979.5  & 66.09  & OOM  \\
        RegenerateImage & N/A & 0\% & 9.53  & 0.1942 & 809.8  & 13.11  & OOM   &  809.8 & 13.11  & OOM   \\
        RegenerateCrop & N/A & 0\% & 54.43  & 0.1737  & 267.3  & 4.30  & 45.00  & 199.2  &  4.19 & 13.66 \\
        \Xhline{3\arrayrulewidth}
        Lazy Diffusion (LD)~\cite{nitzan2024lazy} & 0\% & 0\% & 10.90  & 0.1882  & 1085.1  & 17.13  & 45.00  & 285.8  & 5.68  & 13.66  \\
        LD w/ Model Pruning & 50\% & 30\% &  27.36 & 0.1796 & 775.5   & 15.33  & 45.00   & 204.5  & 5.37  & 13.66  \\
        LD w/ ToMe (TF)~\cite{bolya2023token} & 50\% & 10\% & 60.77  & 0.1756  & 979.0  & 16.87  & 40.14  & 259.7  & 6.82  & 13.01 \\
        LD w/ ToMe (TF)~\cite{bolya2023token} & 50\% & 30\% & 279.36  & 0.1496 & 765.7  & 16.82  & 32.61  & 206.7 & 6.64  & 11.14  \\
        LD w/ AT-EDM (TF)~\cite{wang2024attention} & 50\% & 30\% & 203.35  &  0.1459 & 751.4  & 15.49  & 32.94  & 202.8  & 6.11  & 11.91\\
        \Xhline{3\arrayrulewidth}
        LD w/ \textbf{MoD}  & 50\% & 30\% & 18.34  & 0.1850  & 764.6  & 13.02   & 32.58  & 205.1  & 4.83  & 11.11 \\
        LD w/ \textbf{DiffCR-L} & 50\% & 30\% & 13.53  & 0.1839  & 822.9  & 13.71  & 34.67  & 220.4   & 5.30  & 11.30 \\
        LD w/ \textbf{DiffCR-LT}  & 50\% & 30\% & 13.42  & 0.1845  & 804.9  & 13.89  & 34.88  & 214.5  & 5.21  & 11.52  \\
        \Xhline{3\arrayrulewidth}
        \end{tabular}%
  }
  \vspace{-1em}
  \label{tab:in-painting}%
\end{table*}%

%% file: tabs/hps.tex
\begin{table}[t]
  \centering
  \setlength{\tabcolsep}{3.5pt}
  \renewcommand{\arraystretch}{1.2}
  \caption{Human preference score (HPS) ($\uparrow$) comparison of the proposed DiffCR with other baselines on the T2I task.}
  \label{tab:hps}  %
  \vspace{-0.5em}
  \resizebox{0.8\linewidth}{!}{
        \begin{tabular}{l|c|c}
        \Xhline{3\arrayrulewidth}
        \multicolumn{1}{l|}{\textbf{Methods}} & \multicolumn{1}{c|}{\textbf{DiT C.R.}} & \textbf{HPS Score} \\
        \Xhline{3\arrayrulewidth}
        PixArt-$\Sigma$ (Fine-tuned) & 0\% & 22.582 \\
        PixArt-$\Sigma$ w/ ToMe & 20\% & 16.742 \\
        PixArt-$\Sigma$ w/ MoD & 20\% & 20.580 \\
        PixArt-$\Sigma$ w/ DiffCR & 20\% & 21.427 \\
        \Xhline{3\arrayrulewidth}
        \end{tabular}
  }
  \vspace{-1.em}
  \label{tab:hps}
\end{table}

%% file: sec/5_conclusion.tex
\vspace{-0.3em}
\section{Conclusion}
\label{sec:Conclusion}
\vspace{-0.2em}

In this work, we present DiffCR, a dynamic DiT inference framework with differentiable compression ratios that adaptively routes computation across tokens, layers, and timesteps, resulting in efficient DiT models. Specifically, DiffCR incorporates a token-level routing scheme based on MoD that dynamically learns the importance scores for each token, alongside a novel module that makes MoD differentiable with respect to compression ratios, enabling the model to learn adaptive compression ratios for each layer and timestep. Redundant layers and timesteps learn higher compression ratios, while critical layers and timesteps remain minimally compressed or uncompressed. Extensive experiments on both image inpainting and text-to-image (T2I) tasks consistently demonstrate DiffCR’s superior trade-off between image generation quality and efficiency compared to other compression works.

\section*{Acknowledgment}

The work is supported in part by an internship at Adobe and in part by CoCoSys, one of the seven centers in JUMP 2.0, a Semiconductor Research Corporation (SRC) program sponsored by DARPA.

%% file: sec/6_supple.tex
\clearpage
\appendix
\setcounter{page}{1}
\maketitlesupplementary

\section{More Visualization of Token Routers}
\label{sec:more_visualization}

\begin{figure*}[h]
    \centering
    \includegraphics[width=\linewidth]{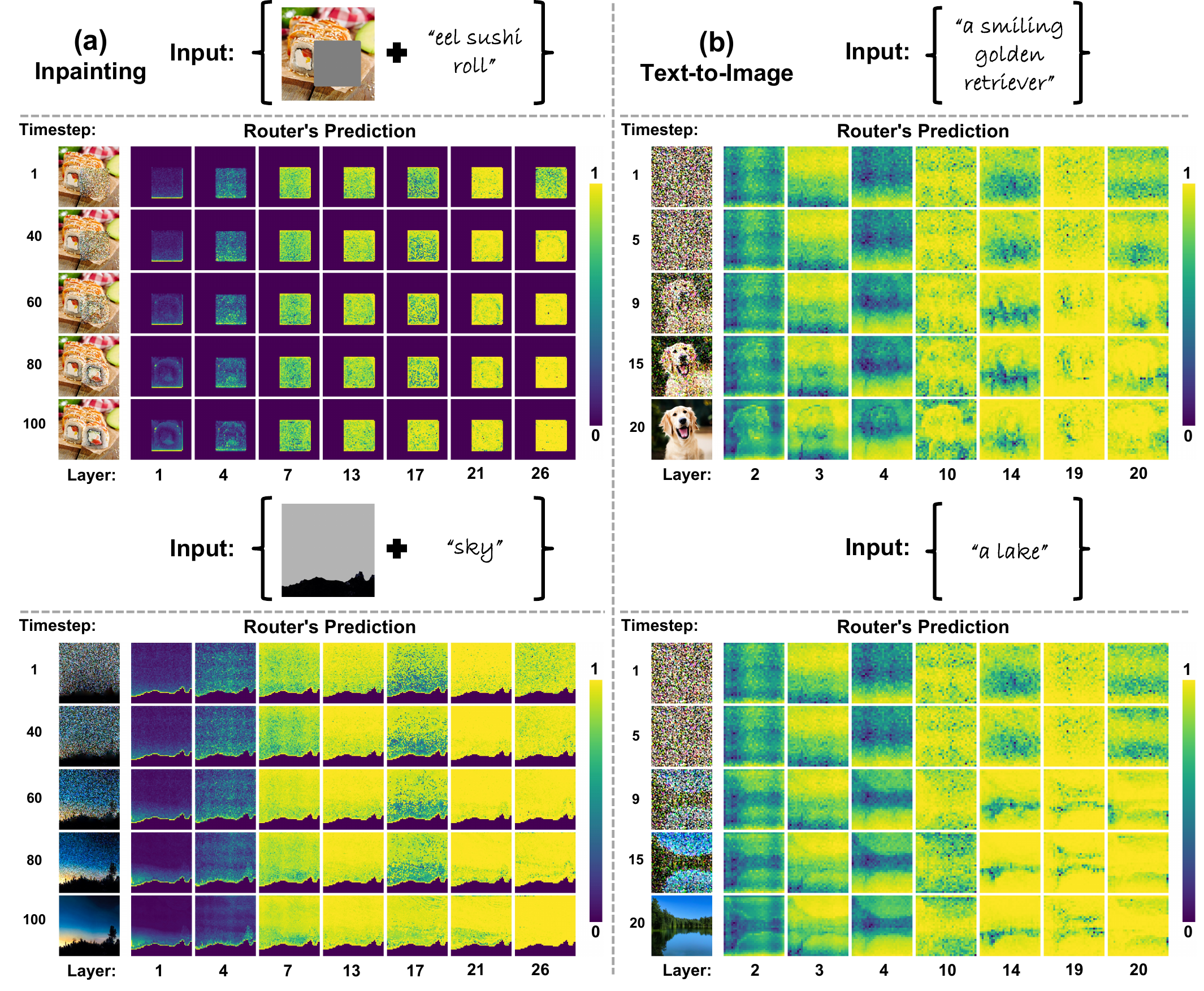}
    \caption{More visualizations of the router's predictions: (a) For inpainting tasks, where inputs are masked images with text prompts, we follow the previous SOTA method Lazy-Diffusion~\cite{nitzan2024lazy} to generate only the masked area rather than the entire image; (b) For text-to-image (T2I) tasks, where inputs are noise and text prompts, we follow PixArt-$\Sigma$~\cite{chen2024pixart} for generation. 
    Each visualization includes the router's prediction map with values ranging from 0 to 1. The generated image at each corresponding timestep is shown on the left, while the router's prediction maps across various layers and timesteps are displayed on the right.}
    \label{fig:more_router_visualization}
\end{figure*}

In Sec.~3.2, we provided an example visualization of the router predictions to evaluate the effectiveness of our DiffCR router. Here, we present additional visualization examples in Fig.~\ref{fig:more_router_visualization} to further validate our findings. 
Our observations consistently demonstrate the following: 
(1) \textit{The router effectively captures semantic information}, clearly delineating object shapes and achieving an attention-like effect while significantly reducing computational costs.  
(2) \textit{The predicted token importance varies across layers and timesteps}. For example, some layers focus on object generation, while others emphasize background areas. Additionally, as timesteps progress, the router increasingly captures the semantic contours of objects, highlighting the importance of dynamic token importance estimation.  
(3) \textit{The optimal compression ratio differs across layers and timesteps}. For instance, some layers assign high importance to all tokens, indicating minimal redundancy, while others selectively prune tokens from objects or backgrounds with distinct shapes, requiring different compression ratios. This variance is also observed across timesteps. In the previous MoD~\cite{raposo2024mixture} approach, a fixed global compression rate is uniformly applied across layers and timesteps, ignoring their individual significance. Such uniform pruning risks over-pruning critical layers or timesteps while under-compressing redundant ones. This observation underscores the need for adaptive and dynamic compression ratios tailored to both layers and timesteps.

\begin{figure}[t]
    \centering
    \includegraphics[width=\linewidth]{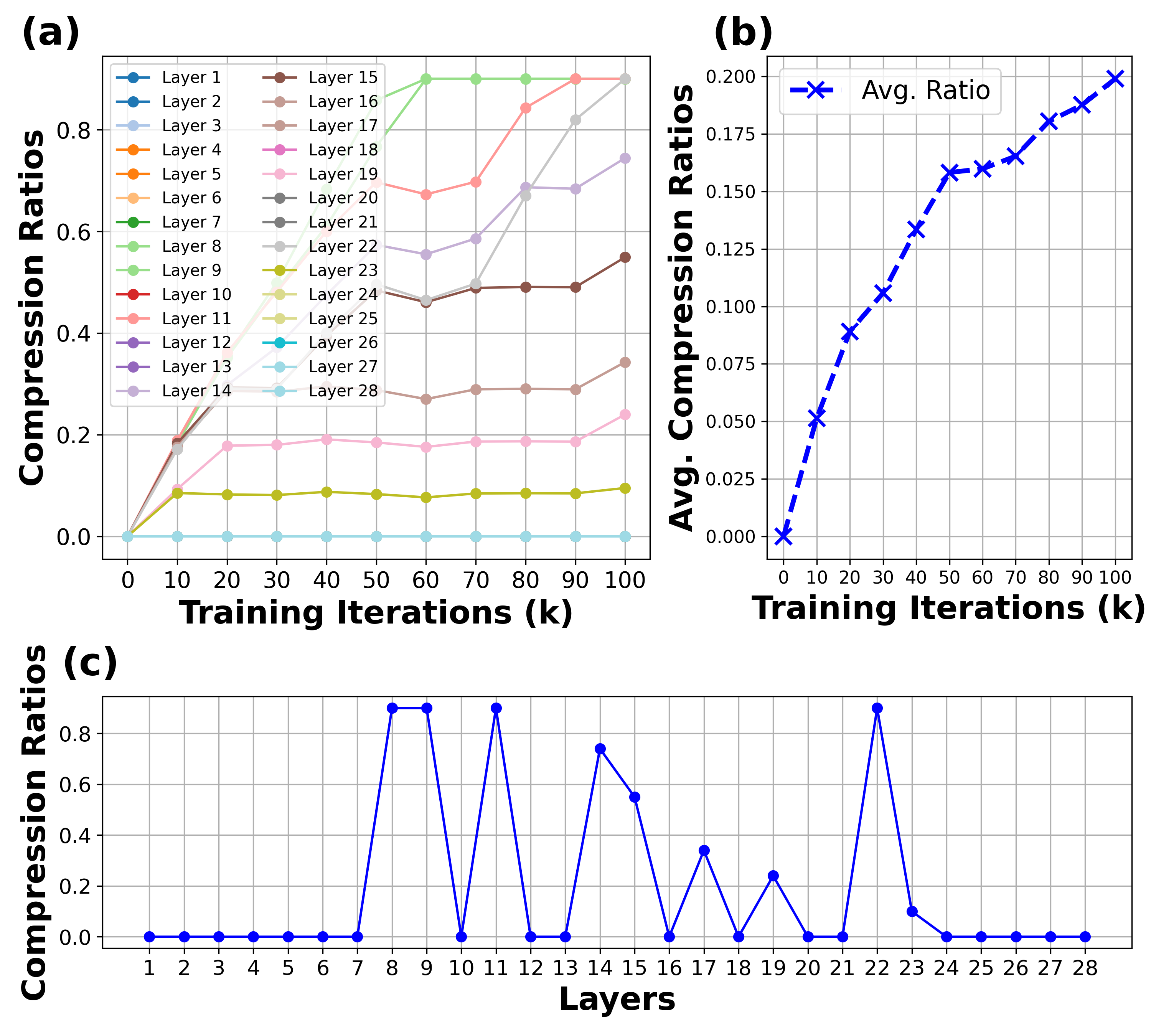}
    \vspace{-2em}
    \caption{Visualization of the 
    compression ratio
    trajectory during fine-tuning for a T2I task: (a) Trajectories for each of the 28 layers in the PixArt-$\Sigma$ model; (b) Average ratio trajectory across all layers; and (c) The final learned ratio distribution across 28 layers.}
    \vspace{-1em}
    \label{fig:layerwise_trajectory_t2i}
\end{figure}

\begin{figure*}[h]
    \centering
    \includegraphics[width=0.65\linewidth]{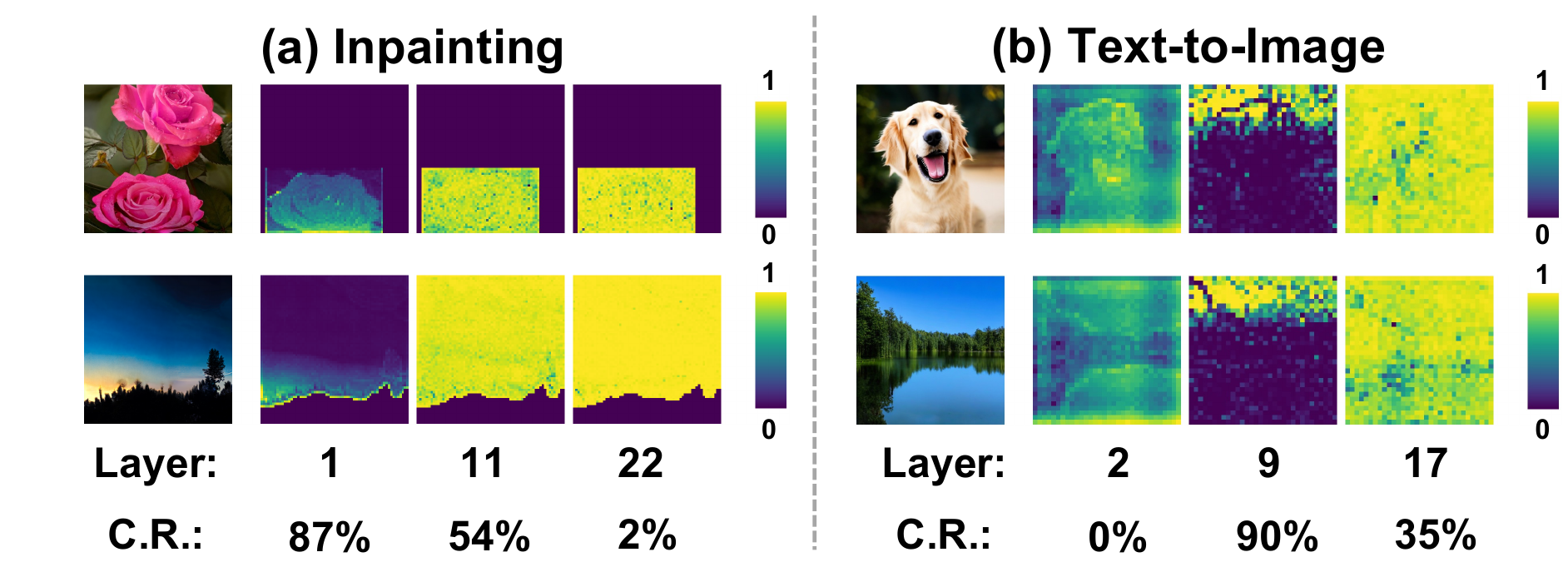}
    \caption{Visualization and analysis of the correlation between the learned compression ratios and the DiffCR router's predictions.}
    \vspace{-1em}
    \label{fig:correlation}
\end{figure*}

\section{Ratio Trajectory Analysis for the T2I Task}
\label{sec:ratio_t2i}

In Sec.~3.3, we visualized the ratio trajectory for inpainting tasks trained with our proposed layer-wise DiffCR. Here, we also provide the training trajectory of compression ratios for all layers during fine-tuning of a PixArt-$\Sigma$ model on a T2I task, as shown in Fig.~\ref{fig:layerwise_trajectory_t2i} (a-c).
The visualization consistently reveals that: 
(1) Each layer learns its unique compression ratio, with redundant layers achieving higher compression and critical layers remaining less or entirely uncompressed;
(2) The average ratio across layers gradually converges to the target ratio. In this example, with a target of 20\%, the final achieved average ratio is approximately 19\%, indicating a minor gap. Notably, a trade-off exists between convergence speed and generation quality: a higher MSE loss coefficient for the ratio accelerates convergence but may degrade quality due to overly rapid compression, while a smaller coefficient promotes gradual convergence and maintains quality, albeit with slower training. 
In practice, we set the initial coefficient to \hr{0.3 and dynamically adjust it during training} to balance speed and quality effectively;
(3) The middle layers exhibit greater redundancy, while the later layers generally have lower redundancy and often cannot be compressed. The early layers show variable redundancy levels.

Note that to prevent the model from learning 0\% compression ratios across all layers, we balance diffusion loss (favoring lower ratios for higher quality) and MSE loss (driving the target average ratio) using a coefficient, without additional regularization or penalties. A higher coefficient speeds up convergence but may compromise quality, while a smaller one ensures gradual convergence and preserves quality.
Some layers naturally learn 0\% ratios, underscoring their importance.

\begin{figure*}[t]
    \centering
    \includegraphics[width=0.8\linewidth]{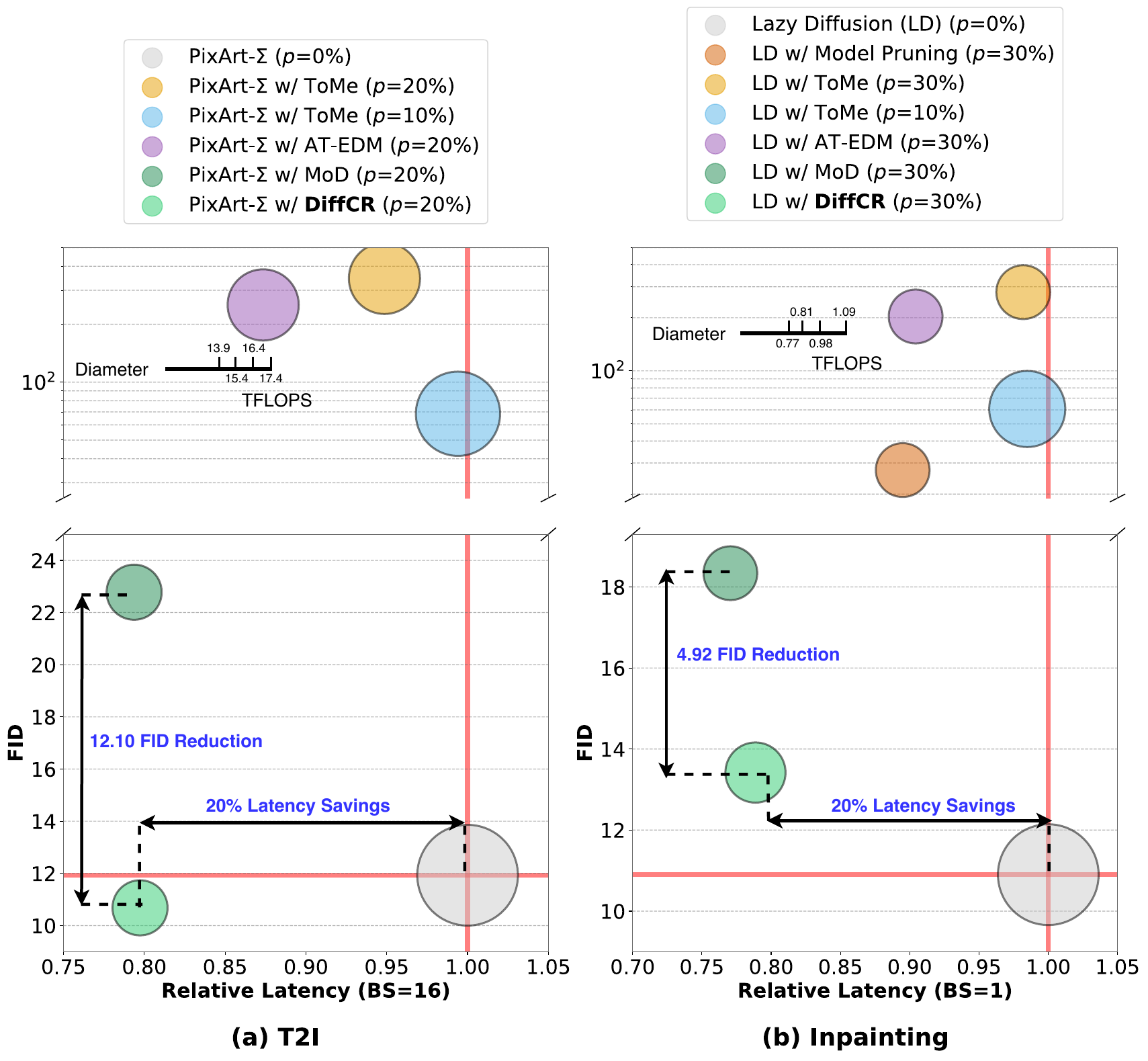}
    \caption{Overall comparison of DiffCR with baselines in terms of latency, FID, and TFLOPS for both T2I and inpainting tasks.}
    \label{fig:overall-comp}
\end{figure*}

\section{Correlation Between Learned Compression Ratios and Router Predictions}
\label{sec:correlation}
We select three representative layers with high, medium, and low learned compression ratios to visualize the corresponding predictions of the DiffCR router and analyze potential correlations. As shown in Fig.~\ref{fig:correlation}, where ``C.R.'' denotes the compression ratios, we observe a strong correlation between the learned ratios and the router's predictions. For layers with high compression ratios, such as layer 1 in inpainting or layer 9 in T2I, the router consistently predicts lower importance scores for many semantic areas, adopting an extremely ``lazy behavior'' to save computations. Conversely, for layers with low compression ratios, the router assigns higher importance scores to most areas. This visualization validates the joint learning effect between our token-level routers and the differentiable ratios.

\section{Trade-offs for Choosing Timestep Regions}
\label{sec:timestep_regions}

In Sec.~3.4, we introduced the timestep-wise DiffCR, where the timestep regions are evenly divided into 10 regions for inpainting tasks with a total of 100 sampling timesteps, and 4 regions for T2I tasks with 20 sampling timesteps. Here, we provide additional guidance on selecting the number of timestep regions and the associated trade-offs.
A larger number of timestep regions allows for learning finer-grained and more precise compression ratios across all timesteps. However, too many regions can make training unstable and challenging. 
To reduce training complexity and enhance stability, we select a smaller number of regions, such as 4 for T2I tasks.
Conversely, using too few regions risks oversimplifying the method, reducing it to heuristic approaches like SpeeD~\cite{wang2024closer}, which manually defines three timestep regions. In practice, we choose between 4 and 10 timestep regions to balance granularity and stability. While our approach aligns with the general insights of SpeeD, it is more systematic and adaptive. Unlike manual exploration of a large design space, our method efficiently handles a significantly greater number of regions in a principled manner, balancing granularity and training stability.

\begin{figure*}
    \centering
    \includegraphics[width=\linewidth]{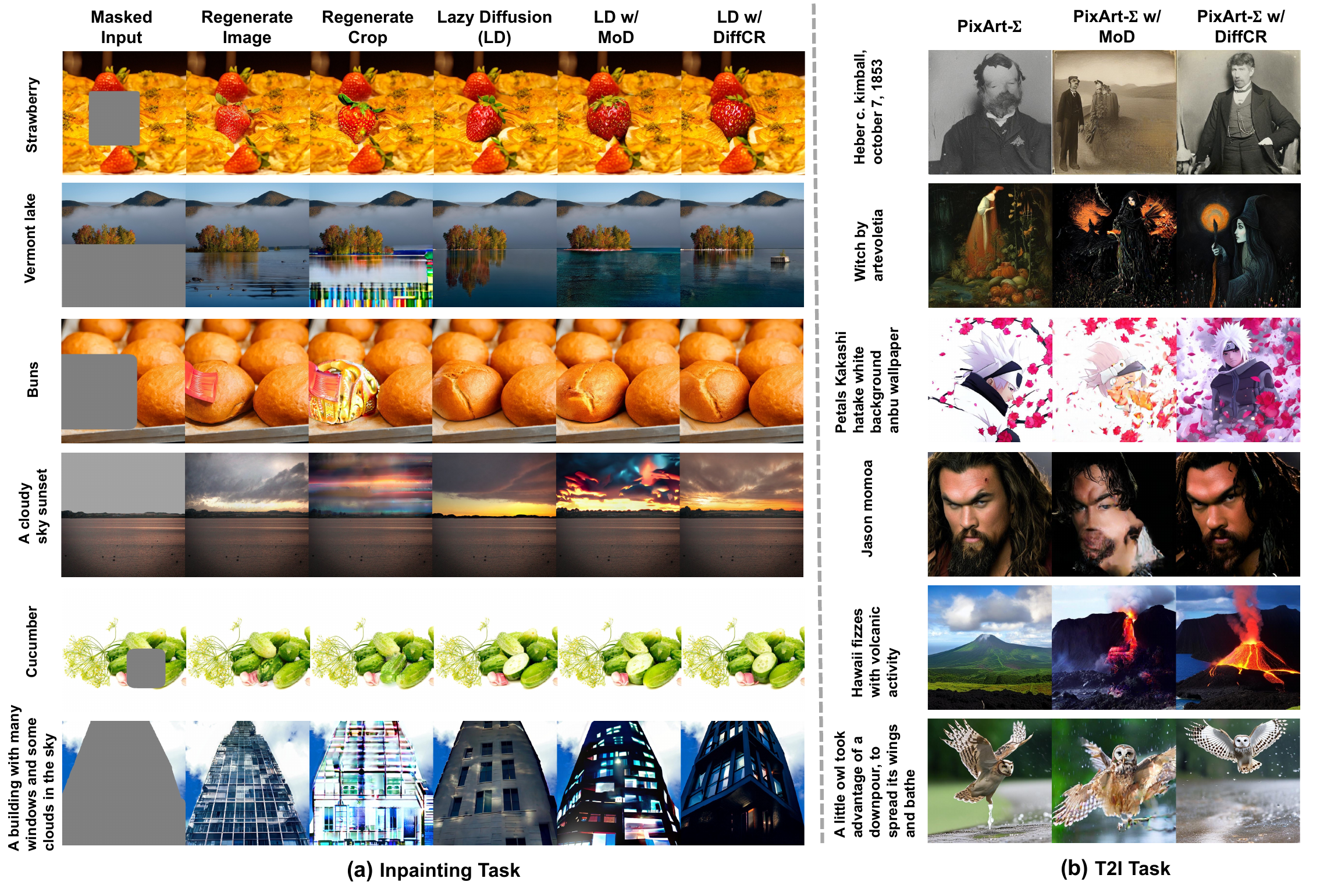}
    \vspace{-1.5em}
    \caption{Additional visual comparisons of our DiffCR with previous uncompressed models and SOTA compression methods: (a) Inpainting tasks, where DiffCR is applied to LD models~\cite{nitzan2024lazy}, and (b) T2I tasks, where DiffCR is applied to PixArt-$\Sigma$~\cite{chen2024pixart}.}
    \label{fig:extra_visual_examples}
    \vspace{-1em}
\end{figure*}

\section{Overall Comparison Figure}
\label{sec:overall_comp}

In Sec.~4.2, we presented a comprehensive comparison of our DiffCR method against baseline approaches for both inpainting and T2I tasks. Here, we provide the overall comparison figures to better illustrate the achieved improvements in FID and latency reductions. 
As shown in Fig.~\ref{fig:overall-comp}, our DiffCR consistently delivers superior trade-offs between FID and latency, achieving FID reductions of 12.10 and 4.92 for T2I and inpainting tasks, respectively, at comparable GPU latency when compared to the most competitive baseline.

\section{Model Trajectories of DiffCR}
\label{sec:more_model_trajectory}

\begin{figure}[t]
    \centering
    \includegraphics[width=\linewidth]{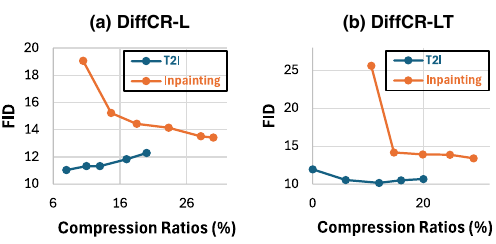}
    \vspace{-1.5em}
    \caption{Model trajectories of DiffCR.
    }
    \label{fig:extra_model_trajectory}
    \vspace{-1em}
\end{figure}

In Sec.~4.2, we visualized the model trajectories during the training of DiffCR-L for both T2I and inpainting 
tasks. This revealed a key benefit: during fine-tuning, the averaged compression ratios across all layers gradually converge to the target ratio, producing a series of ``by-product'' models with varying compression ratios. 
Here, we also supply the model trajectories of DiffCR-LT (``-LT'' denotes layer- and timestep-wise DiffCR). As shown in Fig.~\ref{fig:extra_model_trajectory}, we visualize the FID scores and corresponding compression ratios during the fine-tuning of DiffCR-LT. 
The observations consistently validate the benefits of this approach, showing that it enables the generation of a series of models with diverse compression ratios. Also, we observe that inpainting tasks and Latent Diffusion (LD) models~\cite{nitzan2024lazy} are more sensitive to pruning and require longer fine-tuning to improve generation quality effectively, compared to T2I tasks.
Moreover, for T2I tasks, DiffCR-LT demonstrates slightly greater stability in model trajectory compared to DiffCR-L.

\section{More Visualization of Visual Examples}
\label{sec:more_visual_examples}

\input{tabs/position}

In Sec.~4.4, we selected challenging input prompts to evaluate the qualitative performance of our proposed DiffCR. Here, we provide additional visual examples, as shown in Fig.~\ref{fig:extra_visual_examples}. The examples consistently demonstrate that DiffCR achieves comparable or even superior generation quality compared to the RegenerateCrop baseline and even uncompressed LD or PixArt-$\Sigma$ for inpainting and T2I tasks, respectively. 
Note that ToMe~\cite{bolya2023token} and AT-EDM~\cite{wang2024attention} are omitted here due to their poor generation quality when applied to DiTs, even at a modest compression ratio of 10\%.

\section{Comparison with Caching-based Baselines}  

We summarize the characteristics of our method and caching-based baselines in Tab.~\ref{tab:position}. DeepCache~\cite{xu2018deepcache} and CMYC~\cite{moura2019cache} are designed for U-Net-based models, making direct comparison challenging, while L2C~\cite{ma2024learning} and TGATE~\cite{liu2024faster} target DiTs by caching layer features to reduce recomputation in future timesteps. Unlike these approaches, our method focuses on token pruning with learnable layer- and timestep-dependent compression ratios, and while it does not employ temporal caching, it remains compatible with such techniques. To directly compare, we evaluate all methods using PixArt-$\Sigma$ on the MS-COCO-30K dataset (T2I task) under approximately 25\% latency savings, where L2C achieves an FID of 28.6 (with our trained routers reproducing a similar caching pattern as reported), TGATE yields 43.6 FID, and our DiffCR achieves 28.6 FID. These results show that our method performs comparably to or better than caching-based baselines, and it can be further combined with them to achieve an additional 15 $\sim$ 30\% latency reduction.

\section{Human Preference Score for Inpainting}
\label{sec:hps_inpainting}

In Sec.~4.4, we utilized a computer vision model to estimate likely human preferences and evaluate the ability of models to generate high-quality, contextually relevant images for the T2I task. Here, we also provide the evaluation for inpainting tasks. 
Specifically, we generated 2K samples for the inpainting task and used HPSv2~\cite{wu2023human} to assess human preferences for images produced by different methods. 
As shown in Tab.~\ref{tab:hps_inpainting}, for inpainting tasks, we applied all compression methods to Lazy Diffusion (LD)~\cite{nitzan2024lazy}. DiffCR achieves a higher human preference score of 2.181/0.263 compared to previous compression methods, ToMe~\cite{bolya2023token} and vanilla MoD~\cite{raposo2024mixture}, respectively.

\input{tabs/hps_inpainting}

\section{Ablation Analysis on Compression Ratios}

In this work, we target lower latency as a step toward edge deployment. To analyze the effect of varying compression ratios, we conducted an ablation study using the PixArt-$\Sigma$ model on the MS-COCO-30K dataset~\cite{lin2014microsoft}. Notably, $\nicefrac{1}{3}$ of the timesteps were allocated to full-model inference to preserve accuracy. The results in the table below show that our method scales effectively to larger compression ratios, with only a slight increase in FID $(<1)$. A 30\% compression ratio was previously selected for challenging generation tasks to maintain accuracy while building upon existing state-of-the-art efficient methods.

\input{tabs/ablation}

\section{Is MSE Loss Alone Sufficient?}
We found that simply using the MSE loss effectively guides ratios toward the target without additional regularization, so we fixed it to MSE loss, but other loss functions may also work well. In addition, although we did not enforce binary prediction, the routers tend to learn a polarized distribution in some layers, separating important tokens from unimportant ones, with the learned ratios aligning accordingly, as shown in Fig.~\ref{fig:correlation}.

%% file: tabs/position.tex
\begin{table*}[h]
  \centering
  \setlength{\tabcolsep}{3.5pt}
  \renewcommand{\arraystretch}{1.}
  \caption{Characteristics of our method and caching-based baselines.}
  \vspace{-0.5em}
  \resizebox{0.75\linewidth}{!}{
    \begin{tabular}{l|cccccc}
    \Xhline{3\arrayrulewidth}
    \textbf{Method} & \textbf{Model} & \textbf{Skip / Cache} & \textbf{Granulariy} & \textbf{Learnable} & \textbf{\tabincell{c}{Token \\Pruning}} & \textbf{\tabincell{c}{Timestep-wise \\Feature Cache}} \\
    \Xhline{3\arrayrulewidth}
    DeepCache~\cite{xu2018deepcache} & U-Net & Block & Block & \ding{55} & \ding{55} & \ding{51} \\
    CMYC~\cite{moura2019cache} & U-Net & Block & Block & \ding{55} & \ding{55} & \ding{51} \\
    L2C~\cite{ma2024learning} & DiT & Attn. \& MLP & Layer & \ding{51} & \ding{55} & \ding{51} \\
    TGATE~\cite{liu2024faster} & DiT & Attn. & Layer & \ding{55} & \ding{55} & \ding{51} \\
    \hline
    DiffCR \textbf{(Ours)} & DiT & Attn. \& MLP & Token & \ding{51} & \ding{51} & \ding{55} but compatible \\
    \Xhline{3\arrayrulewidth}
    \end{tabular}%
  }
  \label{tab:position}%
\end{table*}%

%% file: tabs/hps_inpainting.tex
\begin{table}[t]
  \centering
  \setlength{\tabcolsep}{3.5pt}
  \renewcommand{\arraystretch}{1.2}
  \caption{Human Preference Score (HPS) ($\uparrow$) comparison of the proposed DiffCR with baselines for the inpainting task.}
  \label{tab:hps}  %
  \vspace{-0.5em}
  \resizebox{0.75\linewidth}{!}{
        \begin{tabular}{l|c|c}
        \Xhline{3\arrayrulewidth}
        \multicolumn{1}{l|}{\textbf{Methods}} & \multicolumn{1}{c|}{\textbf{DiT C.R.}} & \textbf{HPS Score} \\
        \Xhline{3\arrayrulewidth}
        RegenerateImage & 0\% & 21.056 \\
        RegenerateCrop & 0\% & 19.466 \\
        \Xhline{2\arrayrulewidth}
        Lazy Diffusion (LD) & 0\% & 20.464 \\
        LD w/ ToMe & 30\% &  18.187\\
        LD w/ MoD & 30\% & 20.105 \\
        LD w/ DiffCR & 30\% & 20.368 \\
        \Xhline{3\arrayrulewidth}
        \end{tabular}
  }
  \label{tab:hps_inpainting}
\end{table}

%% file: tabs/ablation.tex
\begin{table}[t]
  \centering
  \setlength{\tabcolsep}{3.5pt}
  \renewcommand{\arraystretch}{1.}
  \caption{Ablation study on the impact of different compression ratios with a batch size of 16.}
  \vspace{-0.5em}
  \resizebox{\linewidth}{!}{
    \begin{tabular}{c|cccccc}
    \Xhline{3\arrayrulewidth}
    \textbf{Metrics$\setminus$Ratios} & \textbf{0\%} & \textbf{10\%} & \textbf{30\%} & \textbf{50\%} & \textbf{70\%} & \textbf{90\%} \\
    \Xhline{3\arrayrulewidth}
    FID Score ($\downarrow$) & 27.80 & 27.53 & 28.64 & 28.57 & 28.44 & 29.21 \\
    CLIP Score ($\uparrow$) & 16.23  & 16.28 & 16.44 & 16.37 & 16.37 & 16.37 \\
    T2I Latency (s) & 11.90 & 11.16 & 10.31 & 9.23 & 8.19 & 7.12 \\
    \Xhline{3\arrayrulewidth}
    \end{tabular}%
  }
  \label{tab:ablation}%
  \vspace{-1em}
\end{table}%